# Contextual Inference in Computational Semantics

Christof Monz

Institute for Logic, Language and Computation (ILLC)
University of Amsterdam, Plantage Muidergracht 24,
1018 TV Amsterdam, The Netherlands
E-mail:                         ,
Phone: +31 20 525 6095, Fax: +31 20 525 5101

**Abstract.** In this paper, an application of automated theorem proving techniques to computational semantics is considered. In order to compute the presuppositions of a natural language discourse, several inference tasks arise. Instead of treating these inferences independently of each other, we show how integrating techniques from formal approaches to context into deduction can help to compute presuppositions more efficiently. Contexts are represented as Discourse Representation Structures and the way they are nested is made explicit. In addition, a tableau calculus is present which keeps track of contextual information, and thereby allows to avoid carrying out redundant inference steps as it happens in approaches that neglect explicit nesting of contexts.

## 1 Introduction

The notion of presupposition has a long tradition in natural language semantics; from early philosophical approaches (e.g., [Str50]) to recent computational approaches (e.g., [PK99]). Almost all accounts of presupposition rely on contextual information in order to compute the presuppositions of a natural language expression. The role that context plays hereby can be seen best by considering an example.

(1) a. Hank likes his wife.
    b. Every man who has a wife likes his wife.

The noun phrase *his wife* behaves as a presupposition trigger which requires that the context in which (1.a) has been uttered provides information allowing to conclude that Hank is married. Compare this to (1.b) where the presupposition trigger *his wife* occurs in the scope of *every* and the restrictor is *man who has a wife*. It is not necessary that the context in which (1.b) is uttered contains the fact that the referent of *his* is married, because this information is provided by the relative clause modifying *man*, where we tacitly assume that the possessive pronoun *his* refers to *a man*. One can say that the relative clause is a local context augmenting the global context in which the whole sentence occurs.

Presupposition triggers are resolved against their local context. If this context provides the presupposed information, then we say that the presupposition does not project. If, on the other hand, the local context does not provide the presupposed information, it does project.

How do we decide whether the context already provides the information expressed by the presupposition trigger? There are basically two ways: [Kar74] states that a presupposition $\pi$ is contained by its context *CON* if *CON* logically entails $\pi$, i.e., if $CON \models \pi$. The second way, as it has been proposed by [vdS92], is to consider presuppositions as anaphoric expressions that have to be resolved against their context. Whereas Karttunen's approach is easy to grasp, we explain van der Sandt's approach in some more detail in the following section. His approach is especially worth considering because, up to now, it seems to be the best approach according to the range of phenomena that can be correctly predicted.

The main goal of this paper is to show how the actual computation of presupposition projection can be improved by combining context and automated deduction. For more general information on context and its use in linguistics the reader is refereed for instance to [Sta98].

One of the few implemented NLP systems which actually compute presupposition projections in natural language discourses is DORIS, cf. [BBKdN99]. We will have a closer look at DORIS later on, and see how formal theories of context, e.g., [AS94a,AS94b], can help devising more efficient and elegant deduction methods for computing presuppositions.

This paper can be regarded as a follow-up of [Mon99], where some of the techniques we are using here have been introduced and applied to a simpler definition of context. Now, we apply some of these techniques to a more complex notion of context employing Discourse Representation Theory (DRT, cf. [KR93]), on which van der Sandt's theory of presupposition projection is based.

The rest of this paper is organized as follows. Section 2 briefly explains van der Sandt's theory of presupposition projection, and shows how it is implemented in the DORIS system. Section 3 introduces a way of extracting the inference problems that arise during the computation of presuppositions and how these problems can be expressed in a fashion that considers the way context is nested within a discourse. In addition, we present a tableau calculus that can be applied to expressions representing contextual information explicitly. Finally, some conclusions and prospects for future work are provided in Section 4.

## 2 Presupposition in DRT

This section provides some background on van der Sandt's approach on treating presuppositions as anaphora. After having introduced the basic data structures, van der Sandt's algorithm for presupposition projection is explained. The second subsection shows how this is realized in the DORIS system and to which extent theorem proving is employed.

## 2.1 Representing Presuppositions as Anaphora

Before we embark on van der Sandt's theory, the concept of an *anaphor* is briefly explained. An anaphor, or anaphoric expression, refers back to something that has been mentioned before. Simple examples are the pronouns *he*, *she*, and *it*. They refer to a (fe)male person, or thing mentioned before, without imposing any further constraints on it. Definite noun phrases containing a possessive pronoun, such as *his wife*, impose more constraints. Here, we are looking for a particular woman mentioned before, who also has to be the wife of a male person who has been mentioned before, too. Obviously, contextual information is necessary to determine the meaning (reference) of an anaphoric expression, and this is also the reason for the strong similarity of presuppositions and anaphora.

Next, we see how anaphoricity is expressed within DRT. DRT is well-suited for explaining anaphora resolution, because it is a dynamic semantics, mainly devised for representing the meaning of discourses, and describing the way contextual information flows through a discourse. The basic data structures of DRT are Discourse Representation Structures (DRSs) which hold the semantic content of sentences as a pair $\langle U, C \rangle$, in which $U$ is a set of variables (or referents) and $C$ is a set of conditions upon them.

**Definition 1 (Discourse Representation Structure).** If $U$ is a set of referents, and $C$ is a set of conditions, then $\langle U, C \rangle$ is a DRS. Let $K_1, K_2$ be DRSs, then K is a condition, if it is of the following form:
$K ::= P(x_1 \ldots x_n) \mid \neg K_1 \mid K_1 \Rightarrow K_2 \mid K_1 \vee K_2 \mid \alpha : K_1$
where $x_1 \ldots x_n$ are discourse referents and $U_1'$ is a subset of the discourse referents of $K_1$.

Alternatively, we will sometimes write DRSs as $[x_1 \ldots x_m | c_1 \ldots c_n]$ because it is less space consuming.

Another important notion within DRT is the accessibility relation which can hold between two DRSs.

**Definition 2 (Accessibility).** A DRS $K_1$ is accessible from a DRS $K_2$ within a DRS $K_0$ if $K_2$ occurs within a condition of $K_1$ or one of the following holds:
$K_1 = K_2$, $K_1 \Rightarrow K_2 \in C_0$
Note, that accessibility is transitive, i.e., if $K_1$ is accessible from $K_2$ and $K_2$ is accessible from $K_3$, then $K_1$ is also accessible from $K_3$.

Due to the limitation of space, we cannot provide any further details on DRT, but the reader is referred to [KR93] for a comprehensive introduction to DRT.

In [vdS92] presupposition triggers are expressed by $\alpha$-DRSs.[1] Consider the sentences in (2) and their respective DRSs in (3).

(2) a. Every man likes his wife.
    b. Every man who has a wife likes his wife.

---

[1] [vdS92] does not call them $\alpha$-DRSs, but here, I follow [Bos94], a very slight modification of van der Sandt's theory, which can be implemented more straightforwardly.

(3)

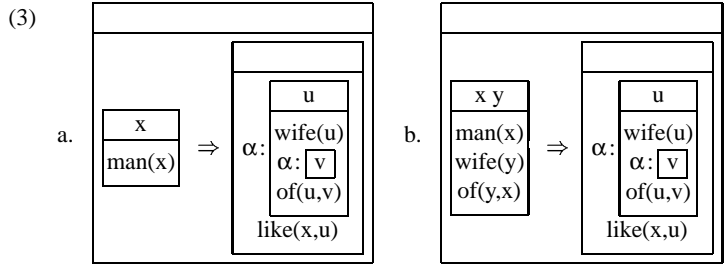

In (3.a), one variable x (or referent) is introduced by the antecedent of the conditional of the DRS. In addition, there is a free variable u to which an α-DRS is attached. This α-DRS contains three conditions. The first condition says that the person u is referring to has to be a wife. α: $\boxed{v}$ is again an anaphor which is referring to somebody in the context, i.e., the accessible DRSs. Finally, both persons have to be in the of-relation.

The α-DRS in (3.b) is exactly the same as in (3.a), but they occur in different contexts. To see whether an α-DRS can be resolved, one has to check whether the DRS that results from substituting the variable which is the argument of the α-operator by an accessible variable is a sub-DRS of the DRS representing the context. The sub-DRS relation is defined as follows:

**Definition 3 (Sub-DRS).** $K_1$ is an immediate sub-DRS of $K_0 = \langle U_0, C_0 \rangle$, if $K_1 = K_0$ or a $c \in C_0$ is of the form:
$\neg K_i$, $K_i \Rightarrow K_j$, $K_i \vee K_j$, or $\alpha : K_i$
where either $i = 1$ or $j = 1$.
The sub-DRS relation is the transitive and reflexive closure of the immediate sub-DRS relation; i.e., if $K_1$ is an immediate sub-DRS of $K_2$, then $K_1$ is a sub-DRS of $K_2$, and, if $K_1$ is a sub-DRS of $K_2$ and $K_2$ is a sub-DRS of $K_3$, then $K_1$ is a sub-DRS of $K_3$.

In the sequel, a context is represented by a DRS. To see whether a DRS $K_1$ occurring in a DRS $K_0$ is entailed by its context, we need an algorithmic way to determine the context of $K_1$ with respect to $K_0$.

**Definition 4 (Context-DRS).** Given a DRS $K_0 = \langle U_0, C_0 \rangle$ and a sub-DRS $K_1$ of $K_0$, the context-DRS of $K_1$ with respect to $K_0$ can be computed recursively:

$$\begin{aligned}
\operatorname{con}(K_1, K_0) &= \langle U_0, C_0 \setminus \{c\} \rangle \oplus \operatorname{con}(K_1, c) \text{ if } K_1 \text{ occurs in } c, c \in C_0 \\
\operatorname{con}(K_1, K_0) &= \langle \emptyset, \emptyset \rangle \text{ if } K_1 = K_0 \\
\operatorname{con}(K_1, c) &= K_2 \oplus \operatorname{con}(K_1, K) \text{ if } c \text{ is of the form:} \\
&\quad K_2 \Rightarrow K_3 \text{, where } K_1 \text{ is a sub-DRS of } K_3 \\
\operatorname{con}(K_1, c) &= \operatorname{con}(K_1, K_3) \text{ if } c \text{ is of the form:} \\
&\quad K_3 \Rightarrow K_2, K_3 \vee K_2, K_2 \vee K_3, \neg K_3 \text{ or } \alpha : K_3 \\
&\quad \text{where } K_1 \text{ is a sub-DRS of } K_3
\end{aligned}$$

In Definition 4 we use the merging function ⊕ (cf. [Zee89]) which allows the information expressed by two DRSs to be merged into one DRS.

**Definition 5 (Merging DRSs).** Given two DRSs $K_1 = \langle U_1, C_1 \rangle$ and $K_2 = \langle U_2, C_2 \rangle$, the merge $K_1 \oplus K_2$ is simply defined as $\langle U_1 \cup U_2, C_1 \cup C_2 \rangle$, the union of the universes and conditions.

Definition 5 assumes that the universes of the DRS are distinct. This means that in general variables are supposed to be introduced only once; i.e., the DRS are *pure*, cf. [KR96]. Although Definition 5 introduces the most simple way of merging, it does not impose any severe restrictions on the expressiveness of the DRS language; but cf. [vEK97] for an overview on different ways of merging.

The resolution tasks of (3.a) and (3.b) are as follows. In (3.a), the only variable that is accessible from v is x, and we see immediately that wife(x) does not occur in the context-DRS of the $\alpha$-DRS. Hence, the $\alpha$-DRS in (3.a) cannot be resolved; i.e., the presupposition projects. In (3.b) on the other hand, substituting v by y allows to wife(y) because it occurs already in the context. Now, substituting u by x allows to solve the whole resolution task as of(y,x) as also part of the context. After resolution, the $\alpha$-DRSs are simply deleted. Because the resolution task in (3.b) can be solved, the presupposition does not project.

In [vdS92], presupposition projection is considered as an instance of accommodation, cf. [Lew79]. Accommodation is a strategy to repair the context in a way such that it allows to conclude the presupposed material. In DRT, repairing the context amounts to adding the presupposed material (the $\alpha$-DRS) to the context. As contexts are represented as DRSs which have themselves internal structure, it is possible to insert an $\alpha$-DRS in several positions in the context. In general, three kinds of accommodation can be distinguished. Considering (3.a), they result in (4).

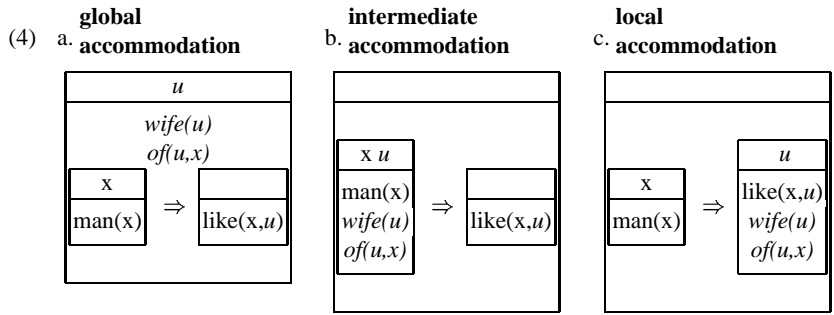

(4) a. **global accommodation**  b. **intermediate accommodation**  c. **local accommodation**

In (4), the accommodated material is typeset in *italics*. Global accommodation adds the presupposed material to the outer-most DRS that is part of the context. Intermediate accommodation adds the $\alpha$-DRS to some DRS that is a proper sub-DRS of the context, but not the DRS in which the $\alpha$-DRS occurs as a condition. Local accommodation simply adds the content of the $\alpha$-DRS to the DRS where the $\alpha$-DRS occurs as a condition.

According to [vdS92], it is only possible to accommodate an $\alpha$-DRS if it contains conditions upon the presupposed variable. Therefore, it is possible to accommodate u and the relevant conditions, but it is not possible to accommodate v, because $\alpha$: $\boxed{v}$ does not contain any further restrictions on v. $\alpha$: $\boxed{v}$ can only be resolved against the context. In (4), $\alpha$: $\boxed{v}$ is resolved to x, the only accessible variable, and v is substituted by x in (4.a)–(4.c).

Deciding which of the different ways of accommodation are correct underlies certain criteria. First of all, accommodation cannot lead to free occurrences of a variable. This constraint is violated by (4.a), where *x* occurs free in *of(u,x)*. In addition, accommodation should preserve local consistency and local informativity, see [Bea97].

**Definition 6 (Local Informativity).** No sub-DRS is redundant. If $K'$ is a sub-DRS of $K$, then $K$ is locally informative if $\text{con}(K', K) \not\models K'$.

**Definition 7 (Local Consistency).** No sub-DRS is inconsistent. If $K'$ is a sub-DRS of $K$, then $K$ is locally consistent if $\text{con}(K', K) \oplus K' \not\models \bot$; i.e., if $\text{con}(K', K) \oplus K'$ is satisfiable.

How the accommodations violating one of those constraints are filtered out in a computational way will be considered in the next subsection.

## 2.2 Implementing Anaphora Resolution

The DORIS system (Discourse Oriented Representation Inference System), cf. [BBKdN99], parses a natural language discourse and generates the corresponding DRS representing its semantic content. This also involves a treatment of presuppositions. Given a sequence of sentences, a DRS possibly containing $\alpha$-DRSs (unresolved presuppositions) is constructed. Then, a generate-and-test procedure returns all DRSs where the presupposed material is either resolved or accommodated and which do not violate the constraints mentioned above.

For instance, if $K_0$ represents a discourse and a sub-DRS $K_2$ of $K_0$ contains an $\alpha$-DRS $\alpha : K_3$ as a condition, then after resolving all simple $\alpha$-DRSs of the form $\alpha : \boxed{x}$ to some accessible variable in $\text{con}(K_3, K_0)$, three possible ways of accommodation are generally possible. In general, local informativity means that the local context of the accommodation site merged with the accommodation site itself do not entail the accommodated DRS. Analogously, local consistency holds if the merge of the local context of the accommodation site and the accommodation site itself and the accommodated DRS is consistent. Global accommodation generates a DRS where $K_3$ is added to $K_0$; i.e., $K_0 \oplus K_3$ is generated. To see whether this obeys local informativity, we have to check whether $\text{con}(K_0, K_0) \oplus K_0 \not\models K_3$ holds. Similarly, being locally consistent means that $\text{con}(K_0, K_0) \oplus K_0 \oplus K_3$ has to be satisfiable. Intermediate accommodation adds $K_3$ to a sub-DRS $K_1$ of $K_0$, which is accessible from $K_3$ and $K_1 \neq K_0$ and $K_1 \neq K_2$. Again, if $K_1 \oplus K_3$ is locally informative and consistent, it has to hold that $\text{con}(K_1, K_0) \oplus K_1 \not\models K_3$, and $\text{con}(K_1, K_0) \oplus K_1 \oplus K_3$ has to be satisfiable. Finally, if $K_3$ is locally accommodated, then it has to be the case that $\text{con}(K_2, K_0) \oplus K_2 \not\models K_3$ holds (locally informative) and that $\text{con}(K_2, K_0) \oplus K_2 \oplus K_3$ is satisfiable (locally consistent). Summing up, we present the six inference tasks that are connected to the different ways of accommodation in Table 1.

Local informativity and consistency can be decided by having run a theorem prover on the different inference tasks.[2] In the sequel, we will focus on local informativity, and the way how it can be computed more efficiently.

Consider example (5.a) and its DRS (5.b).

(5) a. Hank is married. Every man likes his wife.

---

[2] Another and maybe better way of solving the problem of satisfiability is to apply a model generator to the satisfiability task, cf. [BBKdN99].

| | informativity | consistency |
|---|---|---|
| global | $\mathsf{con}(K_0, K_0) \oplus K_0 \not\vdash K_3$ | $\mathsf{con}(K_0, K_0) \oplus K_0 \oplus K_3$ is satisfiable |
| interm. | $\mathsf{con}(K_1, K_0) \oplus K_1 \not\vdash K_3$ | $\mathsf{con}(K_1, K_0) \oplus K_1 \oplus K_3$ is satisfiable |
| local | $\mathsf{con}(K_2, K_0) \oplus K_2 \not\vdash K_3$ | $\mathsf{con}(K_2, K_0) \oplus K_2 \oplus K_3$ is satisfiable |

**Table 1.** Inference tasks for computing informativity and consistency

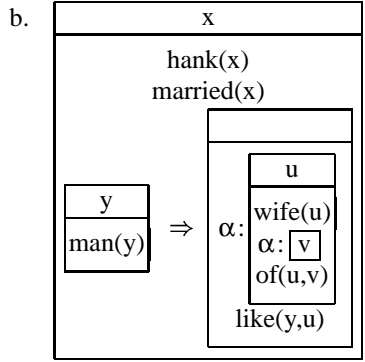

b.

Any straightforward approach to computing the possible accommodation sites has to face five inference tasks, see Table 2, where $\vdash^?$ is an inference task.

If the α-DRS is globally accommodated, only one inference task arises as resolving v to y is ruled out because it violates the free-variable condition. Intermediate and local accommodation have to consider two cases, respectively: one in which v is resolved to x, (ii) and (iv), and the other where v is resolved to y, as in (iii) and (v). DORIS computes the five inference tasks (i)–(v) independently of each other. In this example, the proving method will filter out (i), (iii), and (v), because these are the valid inferences, and thereby they violate local informativity. and (ii) and (iv) remain as possible accommodations sites, since the α-DRSs do not follow from their respective contexts; i.e., they pass the local informativity check.

What is striking about Table 2, is that the DRSs share a lot of information. For instance, the information stemming from the first sentence in (5), namely $\langle \{x\}, \{\mathsf{hank}(x), \mathsf{married}(x)\} \rangle$ occurs in the premise DRS of all five inferences. Therefore, the corresponding deduction rules are applied five times to exactly the same formulas. As far as (5) is concerned, this does not appear to be too dramatic, because this redundancy concerns only the introduction of x and two conditions. On the other hand, in general, the presupposition trigger *his wife* in (5) can occur in a much larger context, containing not only one sentence expressing that Hank is married, but arbitrarily many sentences. In this case, the corresponding DRS representing that context would be much more complex, and consequently the redundancy of treating that context five times would have a much bigger impact on the performance of the computation of the possible accommodation sites: it would slow down significantly.

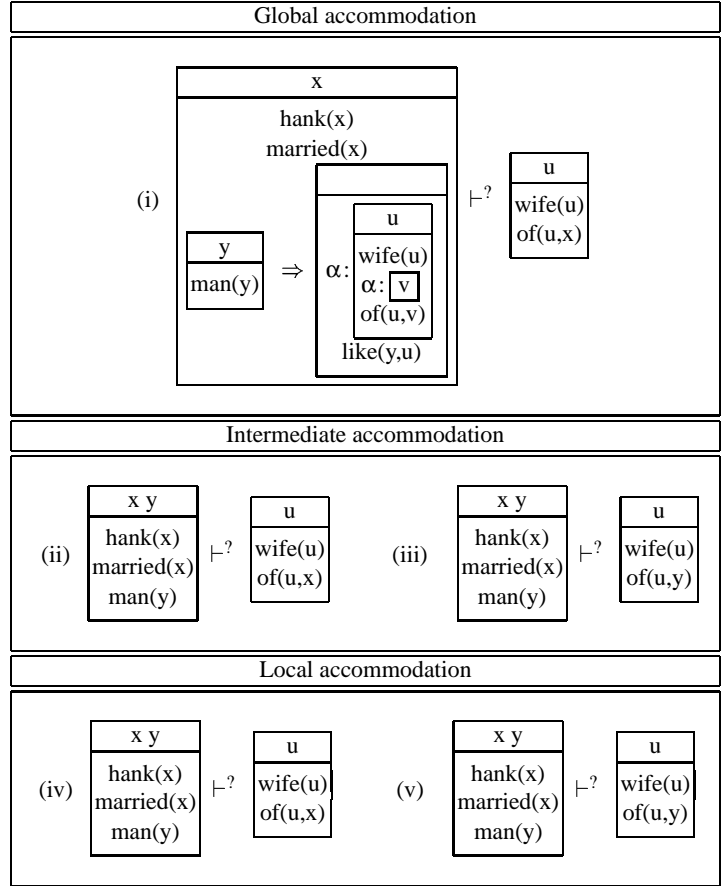

**Table 2.** The inference tasks of (5)

## 3  Integrating Context into Deduction

In the previous section, we have seen that a significant amount of redundancy arises if possible accommodation sites are computed in a straightforward way. This is mainly due to the fact the different inference tasks are treated independently of each other, although they share some information. It is possible to overcome the problem of redundancy by taking context into account. In order to do this, we need a richer language that enables us to express nesting of contexts. Here, we use the in-predicate, cf. [AS94a,AS94b], which takes two arguments. The first argument is a DRS representing the contextual addition, and the second is a conjunction of DRSs and maybe further in-formulas. The second argument represents the consequence that has to hold in the context represented by the first argument. $\mathsf{in}(K_1, \varphi)$ is true if $K_1 \vdash \varphi$.[3] Since $\varphi$ itself can contain an in-predicate, we are able to nest contexts. E.g., $\mathsf{in}(K_1, \varphi \wedge \mathsf{in}(K_2, \psi))$ is true if $K_1 \vdash \varphi$ and $K_1 \oplus K_2 \vdash \psi$. In this case, $K_2$ is a local context for $\psi$. A language containing the in-predicate functions like a meta-language of reasoning, and in the sequel, we give a corresponding tableau calculus. Extending the DRS language with the in predicate results in the language $\mathcal{L}^{con}$.

---

[3] We slightly diverge from [AS94a,AS94b] where the argument positions of in are interchanged.

**Definition 8 (The Language $\mathcal{L}^{con}$).** $\mathcal{L}^{con}$ is defined recursively as follows, where $K_1$ and $K_2$ are DRSs:

$$\varphi ::= P(x_1 \ldots x_n) \mid \neg K_1 \mid K_1 \Rightarrow K_2 \mid K_1 \vee K_2 \mid \mathsf{in}(K_1, \varphi)$$

Note, that $\mathcal{L}^{con}$ does not contain $\alpha$-conditions, as we assume that the formulas of $\mathcal{L}^{con}$ represent possible accommodations. $\mathcal{L}^{con}$ is not used in order to express the semantics of a natural language discourse, but only for expressing which accommodated DRSs have to be evaluated against which context. The purpose of $\mathcal{L}^{con}$ is to express these local informativity problems in a non-redundant fashion. In $\mathcal{L}^{con}$ it is now possible to express the five inference tasks in (2) by a single formula:

(6)  $\mathsf{in}([x|\mathrm{hank}(x), \mathrm{married}(x)], [u|\mathrm{wife}(u), \mathrm{of}(u,x)]$
       $\wedge \mathsf{in}([y|\mathrm{man}(y)], [u|\mathrm{wife}(u), \mathrm{of}(u,x)]$
              $\vee [u|\mathrm{wife}(u), \mathrm{of}(u,y)]))$

The first in-predicate describes the global context as it was relevant for deciding (i) in Table 2. The nested in-predicate augments the global context for deciding whether intermediate and local accommodation are locally informative. In this example, it is not necessary to distinguish between the informativity of intermediate and local accommodation because the DRS which is the local accommodation site does not add any further information to the DRS functioning as the intermediate accommodation site; cf. (ii)–(iv), where the context-DRS remains the same. The disjunction represents the two ways in which the pronouns for intermediate and local accommodation can be resolved; i.e., (ii,iv) vs. (iii,v).

Before a tableau calculus for $\mathcal{L}^{con}$ is presented, it is necessary to show how the inference tasks arising by a DRS $K$ containing $\alpha$-DRSs can be extracted from $K$ and re-stated in $\mathcal{L}^{con}$ in a compact way. We define a function $\tau$ from DRSs to $\mathcal{L}^{con}$. The function $\tau$ is defined in Table 3. As the definition is rather complex and we have only limited space, we just try to sketch its rationale. $\tau$ is recursively applied to a DRS, and it takes two additional parameters: a DRS $K$ representing the current relevant context, and a set $A$ of variables consisting of all accessible variables. It is necessary to keep track of the accessible variables as we have to resolve $\alpha$-DRSs of the form $\alpha: \boxed{x}$, and substitute x by an accessible variable. During the first application of $\tau$ to a DRS, the parameter $K$ is set to $\top$, i.e., the empty or true DRS, and $A = \emptyset$, as no variables have been introduced so far.

The rules in Table 3 are subdivided into three sets of rules. First, if we encounter a DRS which has $\alpha$-DRSs as sub-DRSs, then all the $\alpha$-DRSs are added as conditions of the subordinating DRS. If it is the global DRS, this amounts to global accommodation. The resulting DRS is embedded in an in-predicate if the context is not trivial; i.e., the context-DRS does not equal $\top$. Otherwise, the resulting DRS is not embedded in the in-predicate.

$$[U|C]^{\tau,K,A} = \begin{cases} \mathsf{in}(K, (\bigwedge_{i=1}^{n} \alpha\!:\!K_i^{\tau,\top,A\cup U}) \wedge \langle C, \emptyset \rangle^{\tau,[U],A\cup U}) & \text{if } K \neq \top \\ (\bigwedge_{i=1}^{n} \alpha\!:\!K_i^{\tau,\top,A\cup U}) \wedge \langle C, \emptyset \rangle^{\tau,[U],A\cup U} & \text{if } K = \top \end{cases}$$

$$\langle C \cup \{c\}, C_\alpha \rangle^{\tau,K,A} = \begin{cases} \langle C, C_\alpha \rangle^{\tau,K\oplus[|c],A} & \text{if } c \text{ does not contain an } \alpha\text{-DRS} \\ \langle C, C_\alpha \cup \{c\} \rangle^{\tau,K,A} & \text{if } c \text{ contains an } \alpha\text{-DRS} \end{cases}$$

$$\langle \emptyset, \{c_1,\ldots,c_n\} \rangle^{\tau,K,A} = \begin{cases} \mathsf{in}(K, \bigwedge_{i=1}^{n} c_i^{\tau,\top,A}) & \text{if } K \neq \top \\ \bigwedge_{i=1}^{n} c_i^{\tau,\top,A} & \text{if } K = \top \end{cases}$$

$$\alpha\!:\![U|C]^{\tau,K,A} = \begin{cases} \mathsf{in}(K, \bigvee_{i=1}^{n} [U|C_i]^{\tau,\top,A}) & \text{if } K \neq \top \\ \bigvee_{i=1}^{n} [U|C_i]^{\tau,\top,A} & \text{if } K = \top \end{cases}$$

where $C_1, \ldots, C_n$ are like $C$ but all variables x that occur in an $\alpha$-DRS of the form $\alpha\!:\!\boxed{\text{x}}$ are substitued in $C$ by a variable occurring in A; i.e., the universe of the context. In addition, all $\alpha$-DRSs of the form $\alpha\!:\!\boxed{\text{x}}$ are deleted in $C_1, \ldots, C_n$

$$(\neg[U|C])^{\tau,K,A} = \begin{cases} \mathsf{in}(K, (\bigwedge_{i=1}^{n} \alpha\!:\!K_i^{\tau,\top,A\cup U}) \wedge \langle C, \emptyset \rangle^{\tau,[U],A\cup U}) & \text{if } K \neq \top \\ (\bigwedge_{i=1}^{n} \alpha\!:\!K_i^{\tau,\top,A\cup U}) \wedge \langle C, \emptyset \rangle^{\tau,[U],A\cup U} & \text{if } K = \top \end{cases}$$

$$([U_1|C_1] \Rightarrow K_2)^{\tau,K,A} = [U_1|C_1]^{\tau,K,A} \wedge K_2^{\tau,K\oplus[U_1|C_1],A\cup U_1}$$

$$(K_1 \vee K_2)^{\tau,K,A} = K_1^{\tau,K,A} \wedge K_1^{\tau,K,A}$$

**Table 3.** Extracting informativity tasks from DRSs

The second set of rules sorts conditions which contain $\alpha$-DRSs and those which do not. All conditions which do not contain $\alpha$-DRSs are simply added to the context. Later, the $\alpha$-DRSs will be evaluated against this context.

The last set of rules mirrors how contextual information is threaded through conditions. For instance, the antecedent of an implication is accessible from the succedent, therefore, the antecedent is added to the context parameter of the succedent. Note, that the rules in Table 3 also take care of the free variable constraint, as $\alpha$-DRSs of the form $\alpha\!:\!\boxed{\text{x}}$ are only resolved against the respective contexts of their accommodation sites.

The main advantage of this transformation is that DRS conditions that are used to prove local informativity have to be considered only once. For instance, in (6), it is not necessary to mention again the conditions that are part of the global DRS for checking local informativity of both ways of local accommodation (i.e., whether v is resolved to x or y). In (6), both local accommodation problems are embedded in the global context. Therefore, we can apply the appropriate tableau expansion rule to the $\mathsf{in}$-predicate.

The most important rule of our tableau calculus $\mathcal{T}^{con}$ is the rule $(-\!:\!\mathsf{in})$. Before we introduce the other rules, it is helpful to have a closer look at $(-\!:\!\mathsf{in})$, in order to under-stand the way context is represented in $\mathcal{T}^{con}$.

$$\frac{(i,\sigma,-):\mathsf{in}(K,\varphi)}{\substack{(j,\sigma\cup\{i\},+):K \\ (j,\sigma\cup\{i\},-):\varphi}} \; (-:\mathsf{in})$$

To keep track of the contextual information, labels are attached to the nodes of the tableau. A label has two arguments. Its first argument $i$ is a natural number ($i \in \mathbb{N}$), which is the identifier of the context. I.e., if two nodes have the same number as the first argument of their labels, then they belong to the same context. The second argument $\sigma$ is a set of natural numbers. This set contains the identifiers of the contexts that are accessible. We say that a context $K_1$ is accessible from a formula $\psi$, if there is a formula of the form $\mathsf{in}(K_1,\varphi)$ and $\psi$ is a subformula of $\varphi$. For instance, considering the formula $\mathsf{in}(K_1,\varphi \wedge \mathsf{in}(K_2,\psi))$, $K_1$ is accessible from $\varphi$ and $\mathsf{in}(K_2,\psi)$. Also $K_2$ is accessible from $\psi$. Since accessibility is transitive, it holds that $K_1$ is accessible from $\psi$; but $K_2$ is not accessible from $\varphi$ because $\varphi$ is not embedded in $K_2$ by an $\mathsf{in}$-predicate.

The $(-:\mathsf{in})$-rule is similar to the upwards direction (entering a context) of the (CS)-rule in [BM93]:

$$\frac{\vdash_{\bar{\kappa}*\kappa_1} \varphi}{\vdash_{\bar{\kappa}} \mathsf{ist}(\kappa_1,\varphi)} \; (\mathrm{CS})$$

$\bar{\kappa}$ represents a sequence of contexts and the upwards direction of the rule says that if it is true in the context $\bar{\kappa}$ that $\varphi$ holds in the extension with $\kappa_1$, then $\varphi$ holds in the context $\bar{\kappa}*\kappa_1$ itself. Comparing (CS) to $(-:\mathsf{in})$, we can say that $\bar{\kappa}$ corresponds to $\sigma \cup \{i\}$ and $j$, the identifier of the context extension with $K$ and not $\varphi$, corresponds to $\kappa_1$.

Table 4 gives the complete set of tableau rules. The rules for the usual boolean connectives and quantifiers are omitted, but cf. [Fit96] for a comprehensive introduction to tableau methods.

The contextual information carried by the labels becomes important when we want to define the closure conditions of a branch.

**Definition 9 (Closure of a Branch).** A branch of a tableau tree is closed if it contains two nodes of the form $(i,\sigma,+):R(t_1\ldots t_n)$ and $(j,\sigma',-):R(t'_1\ldots t'_n)$ such that
(a) $t_m$ and $t'_m$ are unifiable ($1 \leq m \leq n$), and
(b) (i) $i = j$ or (ii) $i \in \sigma'$ or (iii) $j \in \sigma$

(a) is the standard condition on branch closure. (b) considers three cases. If $i = j$, then both literals belong to the same context. If $i \in \sigma'$, then $\varphi$ belongs to an extension of $j$. The case where $j \in \sigma$ is analogous to the previous one.

## 4 Conclusions and Future Work

Computing the presuppositions of a natural language discourse is an important task for a natural language processing system. Employing a language like $\mathcal{L}^{con}$ allows for a non-redundant way of stating inference problems that arise in the computation of presuppositions. To this end, we presented a way of extracting local informativity tasks from DRSs

$$\frac{(i,\sigma,+):\mathsf{in}(K,\varphi)}{(j,\sigma\cup\{i\},-):K\,|\,(j,\sigma\cup\{i\},+):\varphi}(+\!:\!\mathsf{in}) \qquad \frac{(i,\sigma,-):\mathsf{in}(K,\varphi)}{\substack{(j,\sigma\cup\{i\},+):K \\ (j,\sigma\cup\{i\},-):\varphi}}(-\!:\!\mathsf{in})$$

$$\frac{(i,\sigma,+):[x_1\ldots x_n|C]}{(i,\sigma,+):[x_1\ldots x_{n-1}|C[x/f(X_1\ldots X_n)]]}(+\!:\!U) \qquad \frac{(i,\sigma,-):[x_1\ldots x_n|C]}{(i,\sigma,-):[x_1\ldots x_{n-1}|C[x/X]]}(-\!:\!U)$$

$$\frac{(i,\sigma,+):[|\{c\}\cup C]}{\substack{(i,\sigma,+):[|C] \\ (i,\sigma,+):c}}(+\!:\!C) \qquad \frac{(i,\sigma,-):[|\{c\}\cup C]}{(i,\sigma,-):[|C]\,|\,(i,\sigma,-):c}(-\!:\!C)$$

$$\frac{(i,\sigma,+):\neg K}{(i,\sigma,-):K}(+\!:\!\neg K) \qquad \frac{(i,\sigma,-):\neg K}{(i,\sigma,+):K}(-\!:\!\neg K)$$

$$\frac{(i,\sigma,+):[x_1\ldots x_n|C_1]\Rightarrow K_2}{(i,\sigma,+):[x_1..x_{n-1}|C_1[x_n/X]]}(+\!:\!\Rightarrow_\forall) \qquad \frac{(i,\sigma,-):[x_1\ldots x_n|C_1]\Rightarrow K_2}{(i,\sigma,+):[x_1..x_{n-1}|C_1[x_n/f(X_1..X_m)]]}(-\!:\!\Rightarrow_\forall)$$

$$\frac{(i,\sigma,+):[|C_1]\Rightarrow K_2}{(i,\sigma,-):[|C_1]\,|\,(i,\sigma,+):K_2}(+\!:\!\Rightarrow) \qquad \frac{(i,\sigma,-):[|C_1]\Rightarrow K_2}{\substack{(i,\sigma,+):[|C_1] \\ (i,\sigma,-):K_2}}(+\!:\!\Rightarrow)$$

$$\frac{(i,\sigma,+):K_1\vee K_2}{(i,\sigma,+):K_1\,|\,(i,\sigma,+):K_2}(+\!:\!\vee) \qquad \frac{(i,\sigma,-):K_1\vee K_2}{\substack{(i,\sigma,-):K_1 \\ (i,\sigma,-):K_2}}(-\!:\!\vee)$$

**Table 4.** The tableau rules of $\mathcal{T}^{con}$

and re-stated them in $\mathcal{L}^{con}$. In addition, a tableau calculus $\mathcal{T}^{con}$ has been presented, allowing to compute informativity problems more efficiently than approaches neglecting context.

Our future work will focus on combining theorem proving and presupposition projection. I.e., whether a presupposition projects or not is only computed if this is necessary in order to derive a certain conclusion. This work is along the lines of [MdR98,MdR99], but it has to deal with more complex data structures representing the context, namely DRSs. In order to to so we have to consider computing local consistency, too; but we think that this can be efficiently accomplished similar to the way local informativity was computed in this paper.

**Acknowledgments.** The author was supported by the Physical Sciences Council with financial support from the Netherlands Organization for Scientific Research (NWO), project 612-13-001.